\documentclass{article}

\usepackage{spconf,amsmath,graphicx}
\usepackage[htt]{hyphenat}
\usepackage{enumitem}
\setlist{nosep, leftmargin=14pt}
\usepackage{hyperref}
\hypersetup{colorlinks=true,linkcolor=[rgb]{0.1,0.3,0.9},urlcolor=[rgb]{0.2,0.2,0.2},citecolor=[rgb]{0.1,0.3,0.9}}

\title{Yucca: A Deep Learning Framework For Medical Image Analysis}

\name{\begin{tabular}{@{}c@{}}
Sebastian Nørgaard Llambias \hspace{0.5em}
Julia Machnio \hspace{0.5em}
Asbjørn Munk \hspace{0.5em}
Jakob Ambsdorf \hspace{0.5em} \\
Mads Nielsen \hspace{0.5em}
Mostafa Mehdipour Ghazi
\end{tabular}}

\address{Pioneer Centre for AI, Department of Computer Science, University of Copenhagen}

\begin{document}

\maketitle

\begin{abstract}

Medical image analysis using deep learning frameworks has advanced healthcare by automating complex tasks, but many existing frameworks lack flexibility, modularity, and user-friendliness. To address these challenges, we introduce Yucca, an open-source AI framework available at \url{https://github.com/Sllambias/yucca}, designed specifically for medical imaging applications and built on PyTorch and PyTorch Lightning. Yucca features a three-tiered architecture: Functional, Modules, and Pipeline, providing a comprehensive and customizable solution. Evaluated across diverse tasks such as cerebral microbleeds detection, white matter hyperintensity segmentation, and hippocampus segmentation, Yucca achieves state-of-the-art results, demonstrating its robustness and versatility. Yucca offers a powerful, flexible, and user-friendly platform for medical image analysis, inviting community contributions to advance its capabilities and impact.

\end{abstract}


\section{Introduction} \label{sec:intro}

The complexity and variability of medical images pose significant challenges for accurate diagnosis and analysis. Traditional image analysis methods often fall short in handling the intricate details and diverse pathologies present in medical data. Advanced Artificial Intelligence (AI) frameworks have become essential tools in medical imaging \cite{ronneberger2015u,zhou2018unet++,lundervold2019overview}, offering sophisticated solutions for tasks such as segmentation, detection, and classification. These frameworks can enhance diagnostic accuracy, streamline workflows, and support large-scale medical research. Despite the availability of several AI frameworks tailored for medical imaging, there is a void of solutions that perform like the popular nnU-Net~\cite{isensee2021nnu} and are flexible like MONAI~\cite{cardoso2022monai}, inhibiting researchers from efficiently developing and deploying customized models for specific medical tasks.

Among the existing AI frameworks, nnU-Net \cite{isensee2021nnu,isensee2024nnu} stands out as the significant advancement in recent times, providing an out-of-the-box solution that competes with state-of-the-art methods. nnU-Net's effectiveness can plausibly be ascribed to four design principles: automatic file management and folder structuring, robust default values, avoidance of common pitfalls, and a curated compilation of best practices. These features enable nnU-Net to adapt seamlessly to a variety of medical imaging tasks with minimal user intervention. However, its highly integrated structure and interdependent modules can be restrictive for users attempting to modify or extend its capabilities.

In contrast, Project MONAI \cite{cardoso2022monai} offers a highly modular and flexible platform for medical imaging research. MONAI provides a comprehensive suite of tools and libraries that support a wide range of applications, emphasizing flexibility and extensibility. This allows researchers to customize the framework to their specific needs. However, the flexibility of MONAI comes at the cost of usability, as the lack of predefined configurations and guidance can pose challenges, especially for users with limited experience in AI and medical imaging.

To address these challenges, we propose Yucca, an open-source, modular, and extendable AI framework designed specifically for medical imaging based on PyTorch \cite{ansel2024pytorch,falcon2019pytorch}. Yucca aims to combine the user-friendly, high-performance characteristics of nnU-Net with the flexibility and modularity of MONAI. By reducing the engineering overhead required to develop and optimize machine learning models, Yucca enables researchers to focus on their specific medical imaging tasks. This makes Yucca accessible to both novice users and experienced researchers, providing a robust and versatile foundation for a wide range of applications. The framework is available on GitHub at \url{https://github.com/Sllambias/yucca}.

\section{Framework Overview} \label{sec:framework}

Yucca is built on a three-tiered design that provides both flexibility and ease of use. We first give a brief conceptual overview of these tiers before detailing each in the following subsections. The first tier, \texttt{Functional}, is inspired by \texttt{torch.nn.functional} and consists solely of stateless functions. This tier shapes the foundational building blocks of the framework, providing essential operations without maintaining any internal state. These functions are designed to be simple and reusable, allowing users to build custom implementations from scratch. The components are modular and can be easily tested and debugged by focusing on pure functions.

The second tier, \texttt{Modules}, is responsible for composing functions established in the \texttt{Functional} tier with logic, and conventions. Modules introduce a layer of structure, handling the organization and processing of inputs and outputs. They encapsulate specific functionalities and are designed to be more user-friendly, reducing the complexity involved in building custom models. While modules rely on more assumptions about the data, they still offer significant flexibility for customization and extension.

The final tier, \texttt{Pipeline}, represents our interpretation of an end-to-end implementation, built upon the previous two tiers. The \texttt{Pipeline} combines modules, functions, heuristics, and decisions into prearranged formulas, akin to the approach taken by the nnU-Net framework. These pipelines are designed to offer complete solutions that require minimal adaptation by the user. Consequently, more behind-the-scenes magic occurs in this tier, with auto-configuration, predefined values, and preselected modules working together to deliver high performance with minimal user intervention. These pipelines prioritize usability and efficiency, to allow users to achieve competitive results without needing to delve into the intricate details of the implementation.

\subsection{Functional}

The \texttt{Functional} tier is primarily populated by micro components that serve as building blocks for macro abstractions. These components range from simple utility operations to more complex functional constructs. The simplest functions in this tier are basic quality-of-life operations, which include tasks such as reading, writing, and converting files of various formats, manipulating paths and directories, and performing basic data integrity tests. These operations serve to provide smooth data handling and management within the framework. Slightly more complex are the array and matrix manipulation functions, which encompass operations such as normalization, filtering, and bounding box calculations. These functions are used to prepare and process image data, to ensure that the data is in the optimal format for subsequent analysis and modeling.

The most sophisticated functions in the \texttt{Functional} tier are compound functions, such as \texttt{preprocess\_case\_for\_inference}. These functions typically accept a significant number of arguments and employ a long sequence of simpler functions to achieve their goals. The compound functions are designed to enable purely functional frameworks to utilize the same processes as the object-oriented Yucca \texttt{Pipeline}. They encapsulate complex logic and workflows into single, reusable entities, facilitating both consistency and ease of use across different parts of the framework. By incorporating these various levels of functions, the \texttt{Functional} tier provides a robust foundation for building custom implementations and supplies users with access to a wide range of tools for data manipulation and preprocessing. This tier's design philosophy emphasizes modularity and reusability, allowing for easy integration and extension within the broader Yucca framework.

\begin{figure*}[!t]
\centering
\includegraphics[width=1\textwidth]{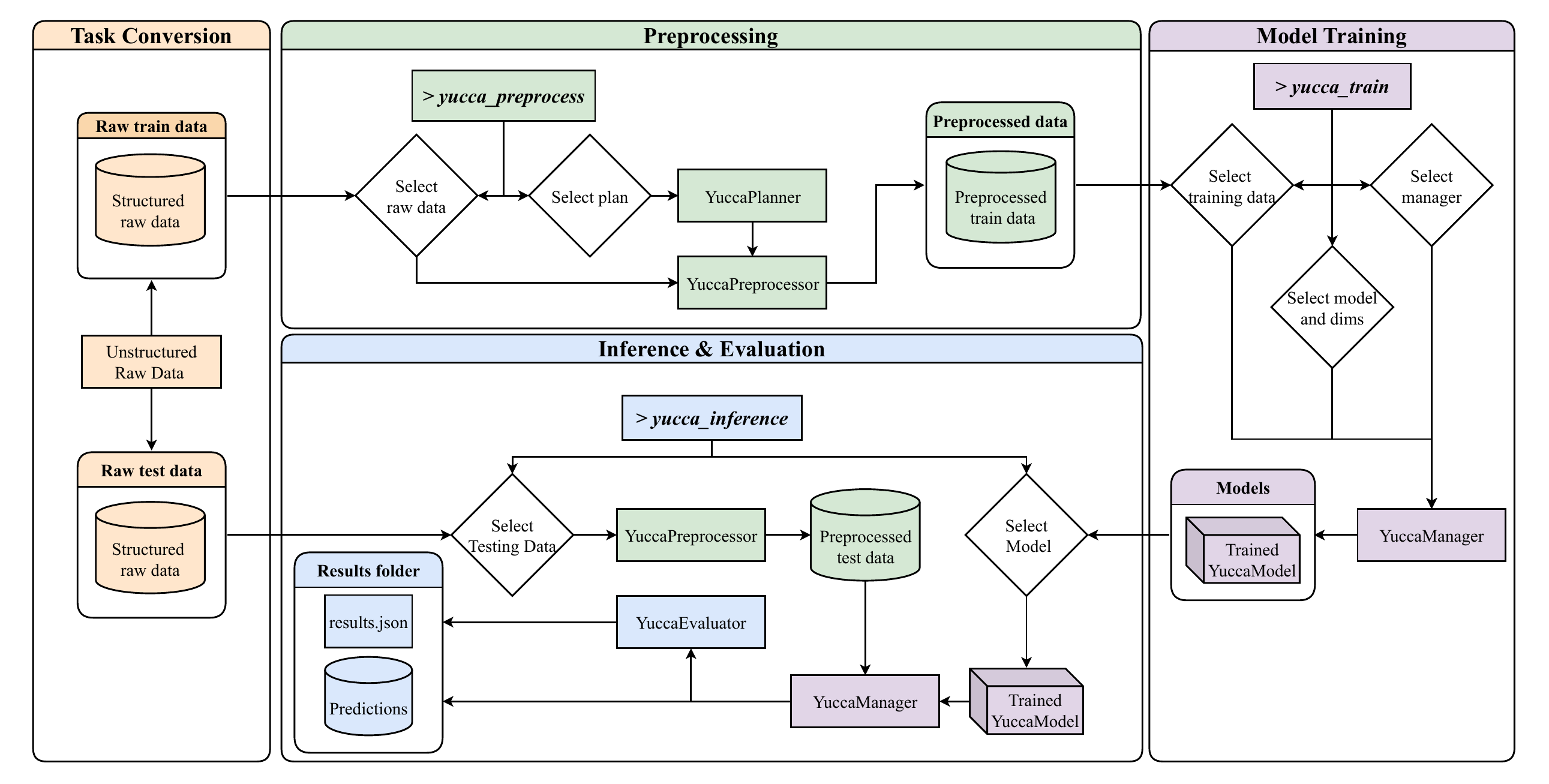}
\caption{The Yucca pipeline consists of four main modules: task conversion, preprocessing, model training, and inference and evaluation. Task conversion structures raw data and splits it into training-validation and testing sets. Preprocessing prepares the training data according to the selected \texttt{YuccaPlanner}. Model training uses the preprocessed data to train the selected model architecture, managed by the \texttt{YuccaManager}. Finally, the inference and evaluation preprocesses the test data, generates predictions, and computes performance metrics using the \texttt{YuccaEvaluator}.}
\label{Yucca_pipeline}
\end{figure*}

\subsection{Modules}

The \texttt{Modules} tier is the first of the two object-oriented tiers of Yucca, where the conventions of the Yucca \texttt{Pipeline} are integrated with the components of the \texttt{Functional} tier. Yucca is built around PyTorch \cite{ansel2024pytorch} and PyTorch Lightning \cite{falcon2019pytorch}, which introduces certain constraints on its module design but also ensures compatibility with other pipelines based on these frameworks. The simplest modules in this tier wrap the micro components of the \texttt{Functional} tier with the logic required by the macro \texttt{Pipeline}. For instance, to use functional transforms with \texttt{Torchvision.Compose}, they are wrapped in callable objects with a defined \texttt{\_\_call\_\_} method. This pattern is similarly applied to functional metrics and loss functions, which are wrapped in callable objects with a defined \texttt{forward} method, aligning with the design of classes inheriting from \texttt{torch.nn.Module}. PyTorch Lightning compliant callbacks, such as the \texttt{WritePredictionsFromLogits} class, also fall into this category. These callbacks utilize specific hooks, such as the \texttt{write\_on\_batch\_end} hook, to execute operations at designated times, like saving predictions during inference using the \texttt{save\_prediction\_from\_logits} function.

More complex modules include the networks, which often combine various block and layer objects along with comprehensive methods to support operations like sliding window inference. These network modules, inheriting from \texttt{torch.nn.Module}, define a \texttt{forward} method that orchestrates the forward pass of the neural network. The most sophisticated modules are the compound \texttt{DataModule} and \texttt{LightningModule} objects required by PyTorch Lightning. These modules encapsulate and streamline much of the code traditionally found in \texttt{train.py} scripts, allowing them to be passed as objects to PyTorch Lightning's \texttt{Trainer}. Briefly, the \texttt{DataModule} organizes and instantiates the training, validation, and inference datasets with the composed transforms, and wraps them in dataloaders with appropriate samplers. The \texttt{LightningModule} manages the training, validation, and inference steps, and defines the optimizers, learning rates, and schedulers. By structuring the framework into these modular components, Yucca provides a robust and flexible environment for building and deploying medical imaging models. The modular design ensures that each component can be individually developed, tested, and reused, facilitating efficient and scalable model development.

\subsection{Pipeline}

The Yucca \texttt{Pipeline} represents the end-to-end implementation of the framework, similar to the nnU-Net framework~\cite{isensee2021nnu}. It combines the tools from the \texttt{Functional} and \texttt{Modules} tiers into a fully operational and user-friendly system. As illustrated in Fig. \ref{Yucca_pipeline}, the pipeline consists of four main modules: (1) task conversion, (2) preprocessing, (3) model training, and (4) inference and evaluation. By integrating these four modules, the Yucca \texttt{Pipeline} offers a comprehensive and streamlined workflow for medical image analysis, enabling researchers to focus on their specific tasks while leveraging the powerful tools and conventions provided by the framework.

The task conversion module handles the initial step of converting unstructured raw data into a structured format that adheres to the Yucca convention. This module also splits the data into training and testing sets. By structuring the data appropriately, the task conversion module sets the stage for efficient and consistent processing in subsequent steps. In the preprocessing module, the training data is prepared according to the specifications of the selected \texttt{YuccaPlanner}. This involves a series of transformations and normalizations designed to enhance the quality of the input data and ensure it is in an optimal format for model training. The preprocessing steps may include tasks such as rescaling, transposition, and intensity normalization, tailored to the specific requirements of the chosen model and the characteristics of the input data.

The model training module is where the core learning process takes place. Users can simply select the desired model architecture, such as a 3D U-Net, and point to the preprocessed training data. The \texttt{YuccaManager} serves as the central engine of this module, orchestrating the training process, managing hyperparameters, and ensuring that the model is trained efficiently and effectively. The final module, i.e., inference and evaluation, handles the process of making predictions on the test data and evaluating the performance of the trained model. The test data is preprocessed using the same pipeline as the training data to ensure consistency. The \texttt{YuccaManager} then generates predictions based on the trained model. Finally, the \texttt{YuccaEvaluator} computes performance metrics by comparing the predictions against the ground truth data, providing a comprehensive assessment of the model's accuracy and robustness.

\subsubsection{Task Conversion}

Task conversion is the process of restructuring a raw, unstructured dataset into the format expected by Yucca. This step includes any data processing that precedes traditional preprocessing, allowing researchers to retrace and reproduce the dataset's processing from its initial manipulation. Examples of such preprocessing steps include excluding corrupt cases, correcting or changing labels, and registering image and segmentation pairs.

The task conversion module also separates the data into training and testing partitions. The subsequent preprocessing and training modules do not have access to files in the test folders, to eliminate any potential data or label leakage during training. This strict separation maintains the integrity of the testing process and the reliability of the model's evaluation.

\subsubsection{Preprocessing}

The Preprocessing module consists of Planners and Preprocessors, which together handle the task of converting and preprocessing raw, task-converted datasets into a standardized format and saving them in the preprocessed folder. First, the user selects a Planner, which analyzes the raw dataset and creates a plan file detailing what preprocessing steps should be taken and how they should be executed. These instructions are derived from both the static specifications of the chosen Planner and the statistical analysis of the raw dataset. For instance, the default \texttt{YuccaPlanner} specifies that images should be resampled to dynamically inferred voxel spacing, whereas the \texttt{YuccaPlanner\_MaxSize} statically resamples images to the dimensions of the dataset's largest image.

Preprocessors then execute the instructions in the plan, with each type of Preprocessor designed to handle specific data configurations. The default \texttt{YuccaPreprocessor} expects image-segmentation pairs as two image files, while the \texttt{ClassificationPreprocessor} expects image-class pairs as an image file and a plain text file. The preprocessing operations applied are limited to those that are consistent each time the sample is drawn, such as normalization, resampling, and transposition. Augmentations, which involve random combinations and magnitudes, are applied online during training and are therefore not included at this stage.

\subsubsection{Training} \label{subsec:training}

\begin{figure*}[!t]
  \centering
  \includegraphics[width=1\textwidth]{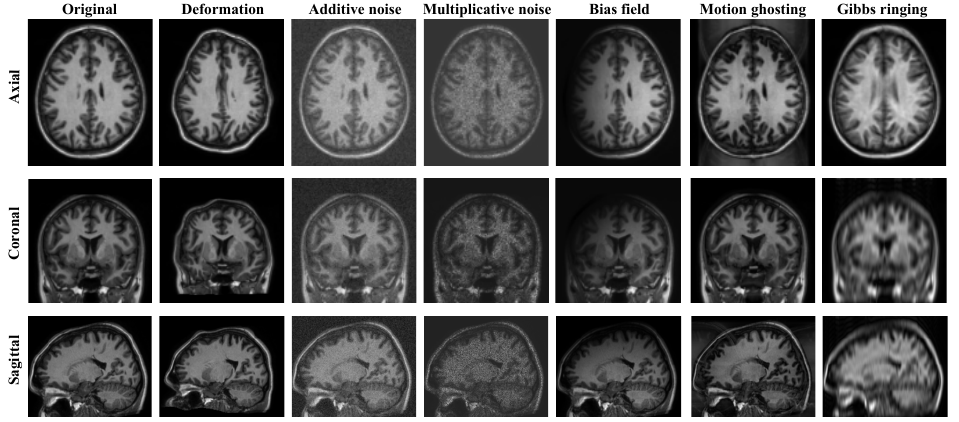}
  \caption{Examples of augmentation techniques provided by Yucca. The first column shows the original T1 image, while the subsequent columns display the image with various augmentation techniques applied. The rows represent different anatomical views in (x, y, z) directions: sagittal (bottom), coronal (middle), and axial (top).}
  \label{augmentation}
\end{figure*}

The Training module is fundamentally responsible for training a model of a specified architecture on a chosen dataset. Central to this module is the concept of a manager, which automates essential but tedious technical aspects and optionally handles critical parameters not manually set by the user. Unspecified settings are selected by the manager based on thoroughly tested heuristics, statistical analysis of the dataset, and the plan created by the \texttt{YuccaPlanner} during preprocessing. This approach mirrors other end-to-end frameworks while also providing users with the flexibility to control every detail of model training, supported by comprehensive documentation and manuals, setting Yucca apart from previous solutions.

When the training module is initialized, the manager autonomously configures settings that, while not directly impacting model performance, are important for reproducibility and proper scientific practice. This includes configuring \textit{multi-channel logging}, \textit{model checkpoints}, \textit{random seeds}, \textit{paths}, and \textit{versioning}. Standardizing the experiment setup helps improve oversight and prevents costly mistakes such as overwriting previous runs or losing track of experiments and results. Paths are configured based on the complete experimental setup and the current version, creating clear and verbose paths. PyTorch callbacks, loggers, and profilers are set up to enable checkpointing, learning rate monitoring, and experiment tracking, with integrated support for Weights and Biases \cite{wandb}.

After configuring the basics, the manager prepares the data. If training and validation \textit{splits} already exist for the task, they are reused to facilitate comparisons between different runs using the same splits. Otherwise, a new split is created and saved in the folder of the preprocessed data. The manager then infers the optimal spatial dimensions based on the data, network architecture, and available hardware. A comprehensive augmentation pipeline is composed \cite{ghazi2022fast}, using many of the augmentations provided by the \texttt{Modules} tier, some of which are shown in Fig. \ref{augmentation}, and the \texttt{DataModule} is instantiated using the previous modules and parameters.

Subsequently, the manager instantiates the network and the optimizers within the \texttt{LightningModule}, automatically resuming unfinished training runs of identical setups. All these modules are then passed to the PyTorch Lightning \texttt{Trainer}. Training is carried out as defined by the \texttt{training\_step} and \texttt{validation\_step} methods of the LightningModule, while the Trainer automates the remaining PyTorch boilerplate code, such as enabling/disabling gradients, calling callback hooks, and moving data to and from devices. Additionally, it provides functionalities like distributed training and gradient accumulation. This approach to training enables researchers to focus on the unique aspects of their models and experiments, while the framework handles the repetitive and technically complex tasks.

\subsubsection{Inference and Evaluation}

The inference module is responsible for making predictions and evaluating the performance of trained models. It takes any trained model and applies it to test datasets that have remained untouched during the training process. First, the test data is automatically preprocessed using the same pipeline used during training. This ensures consistency and limits the distribution shift between the training and test data. The model is then applied to the preprocessed test data, utilizing either full images or sliding window inference for patch-based models.

After predictions are made, any non-label-preserving preprocessing steps, such as transposition or resampling, are reversed. This step is necessary to enable comparison between the predictions and the raw ground truth data. Once the predictions are aligned with the ground truth, they are saved. Finally, the \texttt{YuccaEvaluator} compares the predictions to the ground truth, calculating and saving a comprehensive results file that includes both individual and aggregated metrics. This detailed evaluation provides insights into the model's performance across various metrics, facilitating thorough analysis and comparison.

\section{Results} \label{sec:results}

Yucca has previously demonstrated state-of-the-art performance in various medical imaging tasks, showcasing its robustness in uncontrolled settings. These tasks include but are not limited to cerebral microbleeds segmentation and detection in COVID-19 patients \cite{ferrer2023deep,sagar2024covid}, white matter hyperintensity segmentation and detection in diverse clinical conditions \cite{schiavone2023robust}, hippocampus segmentation \cite{llambias2023data,wu2023active}, and brain lesion segmentation and classification for stroke and multiple sclerosis cases \cite{llambias2024heterogeneous}. These tasks span various 3D and 2D model training scenarios using different deep learning architectures and datasets.

\begin{figure*}[!t]
  \centering
  \includegraphics[width=1\textwidth]{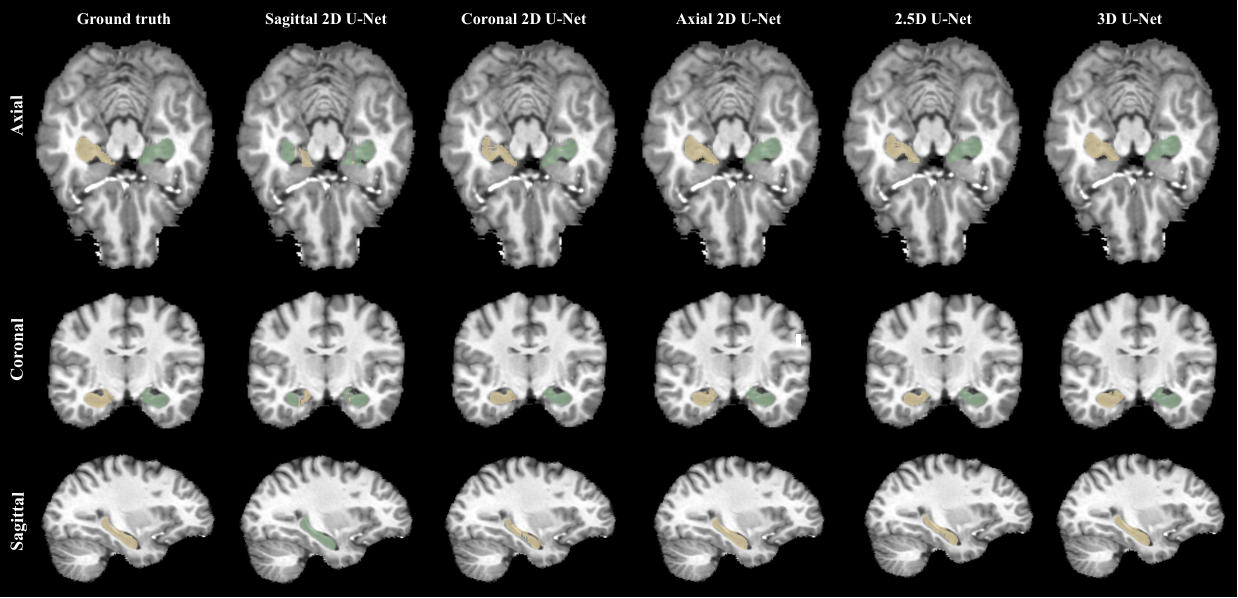}
  \caption{Comparison of the hippocampus segmentation results for a sample T1 MRI using different 2D-3D U-Net architectures.}
  \label{Yucca_results1}
\end{figure*}

To visualize the segmentation ability of the Yucca framework, we experimented with 3D hippocampus segmentation using brain MRIs from the MICCAI Multi-Atlas Challenge \cite{landman2012miccai} dataset. The showcase was conducted using three 2D U-Nets trained on the axial, sagittal, and coronal planes, an ensemble of the three 2D U-Nets, and a 3D U-Net. Qualitative examples are displayed in Fig. \ref{Yucca_results1}, particularly showing the segmentation quality of ensembles and 3D U-Nets.




\section{Conclusion} \label{sec:typestyle}

In this paper, we presented Yucca, a modular and extendable AI framework tailored specifically for medical imaging tasks. Built upon the foundational principles of PyTorch and PyTorch Lightning, Yucca bridges the gap between the robustness of end-to-end solutions like nnU-Net and the flexibility and modularity of frameworks like MONAI. By structuring Yucca into the three tiers of Functional, Modules, and Pipeline, we provide a comprehensive, yet highly adaptable, platform for a wide range of medical imaging applications.

Yucca's design philosophy emphasizes both ease of use and extensibility. The Functional tier comprises stateless, fundamental building blocks, while the Modules tier introduces object-oriented abstractions that integrate seamlessly with the Functional tier. The Pipeline tier offers a complete end-to-end implementation that is easy to use out of the box but remains fully customizable. This layered approach means that Yucca can cater to both novice users seeking straightforward solutions and experienced researchers requiring fine-tuned control over their experiments.

Our evaluation of Yucca across various medical imaging tasks demonstrates its robustness and versatility. Yucca has achieved state-of-the-art results in challenging scenarios such as cerebral microbleeds segmentation and detection in COVID-19 patients, white matter hyperintensity segmentation in diverse clinical conditions, hippocampus segmentation, and brain lesion segmentation and classification for stroke and multiple sclerosis cases. These achievements underscore Yucca's ability to handle 2D and 3D medical imaging tasks with high accuracy and reliability.


In conclusion, Yucca represents a balanced alternative to the existing frameworks in the field of medical imaging AI. Its modular and extendable nature, combined with robust out-of-the-box performance and extensive documentation, makes it an ideal choice for a wide range of users. By providing a solid foundation and the flexibility to customize every aspect of the workflow, Yucca facilitates the development of innovative solutions to complex healthcare challenges in medical imaging. As an open-source project, we invite the community to contribute to Yucca's development and help shape the future of medical imaging research.

\section*{Acknowledgments}

This project has received funding from Innovation Fund Denmark under grant number 1063-00014B, Lundbeck Foundation with reference number R400-2022-617, Digital Research Centre Denmark with grant number 9142-00001B, and Pioneer Centre for AI, Danish National Research Foundation, grant number P1

\bibliographystyle{IEEEbib}
\bibliography{references}

\end{document}